\documentclass[letterpaper, 10pt, conference]{ieeeconf} 

\IEEEoverridecommandlockouts
\usepackage{cite}
\usepackage{amsmath,amssymb,amsfonts}
\usepackage{mathtools}
\usepackage{commath}
\usepackage{algorithmic}
\usepackage{graphicx}
\usepackage{textcomp}
\usepackage{xcolor}
\def\BibTeX{{\rm B\kern-.05em{\sc i\kern-.025em b}\kern-.08em
    T\kern-.1667em\lower.7ex\hbox{E}\kern-.125emX}}

\usepackage{framed}
\definecolor{shadecolor}{rgb}{.9,.9,.9}
\usepackage{todonotes}
\usepackage{siunitx}
\usepackage{booktabs}

\usepackage{tikz}
\usetikzlibrary{quotes,angles}
\usetikzlibrary{calc}
\usepackage{tkz-euclide}\usetkzobj{all}

\usepackage[caption=false, font=footnotesize]{subfig}

\usepackage{times}                 
\usepackage[absolute]{textpos}     
\usepackage[hidelinks]{hyperref}   

\newcommand{\copyrightstatement}{
    \begin{textblock*}{5.7in}(0.25in,0.25in) 

        \noindent
        \footnotesize
        This accepted article to UR 2020 is made available by the authors in compliance with IEEE policy.

        \noindent
        Please find the final, published version in IEEE Xplore.

    \end{textblock*}

    \begin{textblock*}{5.8in}[0,1](0.25in,10.75in) 

        \noindent
        \footnotesize
        \copyright 2020 IEEE. Personal use of this material is permitted.
        Permission from IEEE must be obtained for all other uses, in any current or future media, including reprinting/republishing this material for advertising or promotional purposes, creating new collective works, for resale or redistribution to servers or lists, or reuse of any copyrighted component of this work in other works.
    \end{textblock*}
}

\DeclareMathOperator{\sign}{sgn}

\title{\LARGE \bf
Design and Experiments with a Low-Cost Single-Motor Modular Aquatic Robot
}

\author{Gedaliah Knizhnik and Mark Yim
\thanks{The authors are with the GRASP Laboratory, University of Pensylvannia, Philadelphia, PA 19104. 
        {\tt\small knizhnik@seas.upenn.edu}}%
}

\begin{document}

\copyrightstatement 

\maketitle
\thispagestyle{empty}
\pagestyle{empty}

\begin{abstract}
We present a novel design for a low-cost robotic boat powered by a single actuator, useful for both modular and swarming applications. The boat uses the conservation of angular momentum and passive flippers to convert the motion of a single motor into an adjustable paddling motion for propulsion and steering. We develop design criteria for modularity and swarming and present a prototype implementing these criteria. We identify significant mechanical sensitivities with the presented design, theorize about the cause of the sensitivities, and present an improved design for future work.
\end{abstract}


\section{Introduction}

Simple, inexpensive, and modular robots can provide significant benefits for large scale ocean monitoring and marine operations. Simple designs can be made quiet and non-intrusive to allow for observation of ocean fauna and can be deployed by the hundreds to take measurements over a much wider range than a single robot could cover. Modular robots, meanwhile, can build aquatic structures such as bridges, filters, etc., move them, and disassemble them when done. Such robotic systems could vastly improve our ability to detect and clean oil spills; find, track, and remove ecologically harmful trash collections; or even search for survivors from plane crashes or shipwrecks.

Aquatic robotic teams have been explored previously, but they have generally been limited by cost or locomotion. Leonard et al. considered the use of a team of 10 robotic gliders to explore and sample a coastal region \cite{Leonard2007}\cite{Leonard2010}, but the use of gliding locomotion restricts the permitted task space. Mintchev et al. presented a design for a miniature AUV capable of swarming behavior and full 3D motion \cite{Mintchev2014}, but it involves multiple complex, custom, and expensive actuators. Modular aquatic robots were explored in the Tactically Expandable Maritime Platform (TEMP) project, which deployed 33 modules to autonomously construct an aquatic bridge \cite{OHara2014}\cite{Paulos2015}, but required complex holonomic actuation. Furno et al. also presented a design for a modular underwater robot capable of docking and reconfiguration \cite{Furno2017}, as did Mintchev et al. \cite{Mintchev2014Journal}, but similar work is limited. 

Actuation thus presents a barrier to the development of significant modular and swarming aquatic systems, and a simple, inexpensive module is needed to propel development. Propulsion via an internal rotor and the conservation of momentum is a novel and inexpensive mechanism that can enable this work. Originally explored in the context of terrestrial locomotion, such as by Kelly et. al with the Chaplygin Sleigh \cite{Kelly2012} and Degani for vertical climbing between two walls \cite{Degani2010}\cite{Degani2016}, it has recently been applied to aquatic locomotion as well. A fishlike robot based on the design was developed in \cite{Kelly2012}, and Tallapragada showed that the core propulsion mechanism was vortex shedding from the tail due to the internal rotor\cite{Tallapragada2015}.

A novel approach to the internal rotor in an aquatic environment was presented by Refael and Degani, who explored using an internal rotor to induce a paddling motion in a set of passive flippers by the conservation of angular momentum \cite{Refael2015}\cite{Refael2018}, but their design is hampered by limited mobility. In this paper we extend the work of Refael and Degani, presenting and characterizing a \textbf{new design} for a momentum-driven swimming robot capable of functioning as a mobile sensor platform individually or \textbf{in a coordinated or modular group} to overcome its limited mobility.

\begin{figure}[t]
    \centering
    \includegraphics[width=\linewidth]{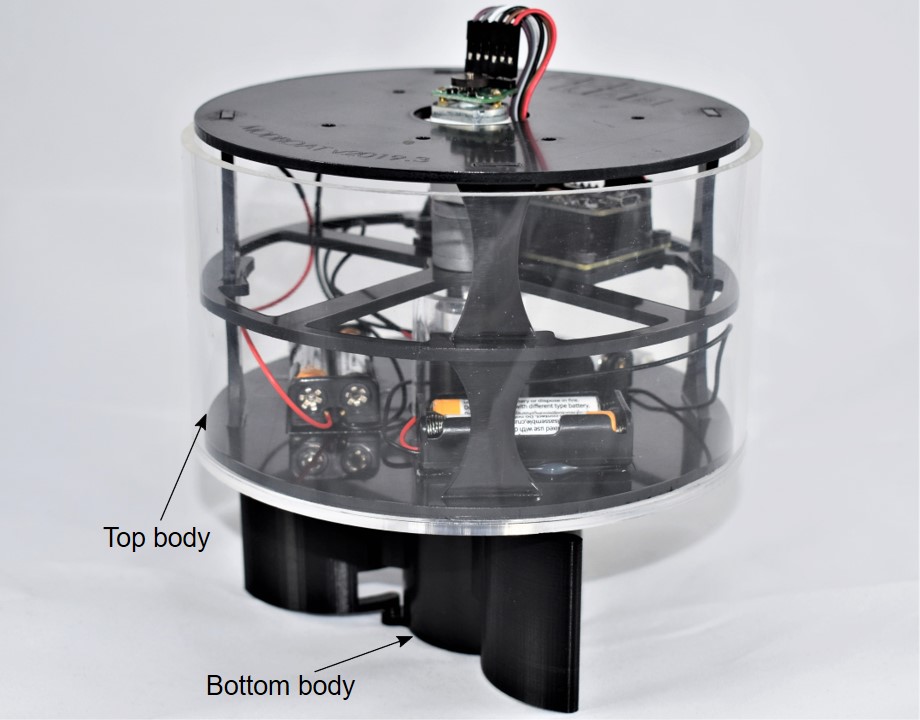}
    \caption{The Modboat prototype, which is described in Section \ref{sec:mechDes}.}
    \label{fig:renderingIso} 
\end{figure}

This paper is organized as follows: in Section \ref{sec:basicDes} we present the basic design developed in \cite{Refael2018} and equations of motion for it. In Section \ref{sec:swarmAndModul} we discuss the design principles that we apply to the basic model to allow modularity and swarming behavior, and in Section \ref{sec:mechDes} we introduce our design: the Modboat. In Section \ref{sec:experiments} we present experiments done to evaluate the design; we discuss the results in Section \ref{sec:discussion} and offer an improved design.


\section{Design Principle and Modeling} \label{sec:basicDes}

\subsection{Dynamic Model} \label{sec:model}

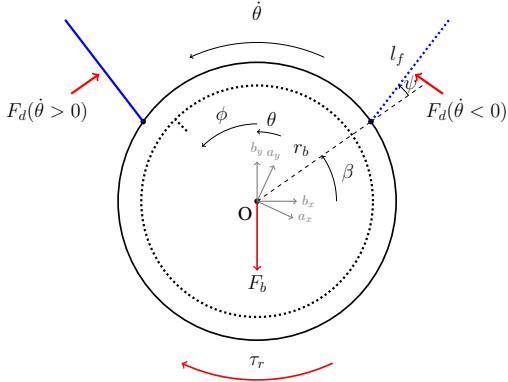
\begin{figure}[t]
    \centering
    \resizebox{!}{0.6\linewidth}{%
    \begin{tikzpicture}
        \def\radius{3.5}
        \coordinate (O) at (10,10);
        \path (O) +(145:\radius) coordinate (FL);
        \path (O) +(35:\radius) coordinate (FR);
        \path (O) +(35:1.5*\radius) coordinate(FRRef);
        
        \draw[very thick] (O) circle[radius=\radius];
        \fill (O) circle[radius=2pt] node[below left] {\Large O};
        \fill (FL) circle[radius=2pt];
        \fill (FR) circle[radius=2pt];
        
        \begin{scope}[rotate around={10:(FL)}]
        \path (FL) arc (0:55:\radius) coordinate (FLEnd);
        \end{scope}
        \draw[ultra thick, blue] (FL) -- (FLEnd);
        
        \begin{scope}[rotate around={170:(FR)}]
        \path (FR) arc(0:-55:\radius) coordinate (FREnd);
        \end{scope}
        \draw[ultra thick, blue, dotted] (FR) -- (FREnd);
        \node[above left](FRMid) at ($(FR)!0.5!(FREnd)$) {\Large $l_f$};
        
        \draw[dashed] (O) -- (FR);
        \draw[dashed] (FR) -- (FRRef);
        \node[above left](RRad) at ($(O)!0.5!(FR)$) {\Large $r_b$};
        
        \coordinate(AX) at (O);
        \draw[gray, thick, ->] (AX) --++ (0,1) node[above] (YA) {$b_y$};
        \draw[gray, thick, ->] (AX) --++ (1,0) node[right] (XA) {$b_x$};
        
        \begin{scope}[rotate around={-25:(O)}]
        \draw[gray, thick, ->] (AX) --++ (0,1) node[above] (YB) {$a_y$};
        \draw[gray, thick, ->] (AX) --++ (1,0) node[right] (XB) {$a_x$};
        \end{scope}
        
        \draw
            pic["\Large $\beta$",draw=black, ->, angle eccentricity=1.2, angle radius=2.0cm]{angle=XA--O--FR};
        
        \draw
            pic["\Large $\psi$",draw=black, ->, angle eccentricity=1.2, angle radius=1.15cm]{angle=FRRef--FR--FREnd};
            
        \draw
            pic["\Large $\theta$",draw=black, ->, angle eccentricity=1.2, angle radius=1.75cm]{angle=YB--O--YA};
            
        \path (O) +(65:1.5*\radius) coordinate (RTD);
        \path (O) +(115:1.5*\radius) coordinate(LTD);
        \draw
            pic["\Large  $\dot{\theta}$",draw=black, ->, angle eccentricity=1.2, angle radius=4cm]{angle=RTD--O--LTD};
        
        \path (O) +(270:\radius/2) node[below](BD) {\Large $F_{b}$};
        \draw[very thick, red, ->] (O) -- (BD);
        
        \path (O) +(245:1.25*\radius) coordinate (RFR);
        \path (O) +(295:1.25*\radius) coordinate (LFR);
        \draw
            pic["\Large $\tau_r$",draw=red, very thick, <-, angle eccentricity=.9, angle radius=4.5cm]{angle=RFR--O--LFR};

        \node[](FRMid1) at ($(FR)!0.5!(FREnd)$) {};  
        \begin{scope}[rotate around={-125:(FRMid1)}]
        \path (FRMid1) +(90:\radius/2) node (FDEnd) {\Large $F_d(\dot{\theta} < 0)$};
        \end{scope}
        \draw[ultra thick, red, <-] (FRMid1) -- (FDEnd);
        
        \node[](FLMid1) at ($(FL)!0.5!(FLEnd)$) {};  
        \begin{scope}[rotate around={125:(FLMid1)}]
        \path (FLMid1) +(90:\radius/2) node (FDEnd1) {\Large $F_d(\dot{\theta} > 0)$};
        \end{scope}
        \draw[ultra thick, red, <-] (FLMid1) -- (FDEnd1);
        
        \draw[ultra thick, dotted] (O) circle[radius=\radius/1.2];
        \path (O) +(135:\radius/1.4) coordinate (X1);
        \path (O) +(135:\radius/1.2) coordinate (X2);
        \draw[ultra thick, dotted] (X1) -- (X2);
        
        \draw
            pic["\Large $\phi$",draw=black, ->, angle eccentricity=1.2, angle radius=1.95cm]{angle=YA--O--X2};
        
    \end{tikzpicture}}
    \caption{A functional diagram of the robot design. The motor is mounted at $O$, and the orientation of the driving mass is given by $\phi$. $\theta$ defines the orientation of the body-fixed frame $b$ in the world frame $a$. Although both flippers are shown as fully open, only one is open at any given time, as indicated by the solid line. Forces and torques considered are shown in red.}
    \label{fig:diagramSimple} 
\end{figure}

We consider a single-actuated swimming robot based on the design presented by Refael and Degani  \cite{Refael2015}, \cite{Refael2018}. It consists of a single cylindrical body, while a motor in the center rotates a driving mass --- a symmetric mass with a high moment of inertia. Two flippers are mounted to the underside of the body and allowed to rotate freely, with hard stops defining a closed position and a fully open position. A representative diagram is shown in Fig. \ref{fig:diagramSimple}, where both flippers are shown open for clarity.

The robot moves by using the conservation of angular momentum. When the motor spins the driving mass, conservation of momentum requires that the body rotate in the opposite direction, with the amount of rotation proportional to the relative inertias of the body and the mass. The angular acceleration this creates, as well as drag from the fluid, causes the leading flipper to open against its hard stop, as shown in Fig. \ref{fig:diagramSimple}. At this point the drag ($F_d$) acts as thrust, pushing the robot laterally and forward. If the motor rotation is reversed, the open flipper is pulled closed, while the new leading flipper is opened. The lateral thrust elements cancel out, while the forward motion is preserved, resulting in a net forward motion.

We define a model to analyze the motion, in which we let $\theta$ define the orientation of the body in an inertial frame, while $\phi$ defines the angle of the driving mass relative to the body. Finally, let $(x,y)$ be the position of the center of the robot in the inertial frame. Then the configuration of the system is $\begin{bmatrix} x & y & \theta \end{bmatrix}^T$. Assuming that we have sufficient control of the motor, we take $\phi(t)$ as the prescribed input variable.


To simplify the model of the robot, we make the following assumptions:

\begin{enumerate}
    \item The flippers open and close instantaneously and provide thrust only when fully open.
    \item Only one flipper is open at any time, determined by $\sign{(\dot{\theta})}$. 
    \item The linear velocity of the robot is negligible compared to the rotational velocity. Thus the force on the flaps depends exclusively on $\dot{\theta}$ and not on $\dot{x}$ or $\dot{y}$. 
    \item The fluid velocity across each flap is approximately constant and equal to the fluid velocity at the center of the flap ($\norm{v_{f}}$). 
    \item The velocity of the water everywhere else is $0$; i.e. there are no external flows.
\end{enumerate}

The following forces, shown in Fig. \ref{fig:diagramSimple}, are considered to develop the equations of motion:

\begin{enumerate}
    \item Drag on the open flipper $F_d$, which is modeled as a flat plate. The closed flipper is ignored. 
    \item Linear drag $F_b$ on the robot body as it moves through the fluid. 
    \item Rotational drag $\tau_r$ on the robot body as it rotates within the fluid (this is not considered in \cite{Refael2018}). 
\end{enumerate}

\begin{table}[t]
    \centering
    \caption{Parameters in \eqref{eq:eom_main} and \eqref{eq:variablesMain}. Values provided are for the design presented in Section \ref{sec:mechDes}.}
    \begin{tabular}{llrl} \toprule
    Var. & Description & Value & Units \\ \midrule
        $m$     & Total mass of the robot  & 0.63 & $\si{kg}$ \\
        $I_{t}$ & Inertia of driving mass & \num{1.6e-3}& $\si{\kg \square\metre}$ \\
        $I$ & Inertia of whole robot & \num{1.7e-3} & $\si{\kg \square\metre}$ \\
        $A_{sub}$ & Submerged body area & 0.025 & $\si{\square\metre}$ \\         
        $r_{t}$  & Driving mass radius                   & 0.075 & $\si{\metre}$ \\
        $r_{b}$  & Bottom body radius                   & 0.025 & $\si{\metre}$ \\
        $l_{f}$  & Flipper length                        & 0.050 & $\si{\metre}$ \\
        $d_{f}$  & Flipper submerged depth               & 0.043 & $\si{\metre}$ \\
        $\beta$  & Flipper angular location              & 45 & $\si{\degree}$\\
        $\psi$   & Flipper maximum open angle            & -21 & $\si{\degree}$\\
        $\rho$   & Density of water                   & 1000 & $\si{\kg\per\cubic\metre}$ \\ 
        $C_b$ & Body translation drag coeff. & 1.0 & --- \\
        $C_r$ & Body rotation drag coeff. & 1.2 & ---       \\
        $C_f$ & Flipper drag coeff. & --- & --- \\ \bottomrule
    \end{tabular}
    \label{tab:params}
\end{table}

The full derivation and a partial analysis of the assumptions can be found in \cite{Refael2018}, whose model we have reproduced for clarity. The final equations of motion are given in \eqref{eq:eom_main}, with coefficients given by \eqref{eq:variablesMain}.

\begin{align} 
    M \begin{bmatrix} \ddot{x} \\ \ddot{y} \\ \ddot{\theta} \end{bmatrix} =& \dot{\theta}^2 K_{f} R \begin{bmatrix} \sin(\beta + \psi)\sign(\dot{\theta})  \\ \cos(\beta + \psi) \\ - K_{t} \sign(\dot{\theta})  \end{bmatrix} -  K_{b} \norm{v} \begin{bmatrix} \dot{x} \\ \dot{y}  \\ 0 \end{bmatrix} \nonumber \\
        &   - \begin{bmatrix} 0 \\ 0 \\ C_r \dot{\theta} \end{bmatrix} -\begin{bmatrix} 0 \\ 0 \\ I_{t} \ddot{\phi} \end{bmatrix} \label{eq:eom_main}
\end{align}

\begin{subequations} \label{eq:variablesMain}
\begin{align}
M &=  \begin{bmatrix} m & 0 & 0 \\ 0 & m & 0 \\ 0 & 0 & I \end{bmatrix} \\
R &= \begin{bmatrix} \cos(\theta) & -\sin(\theta) & 0 \\ \sin(\theta) & \hphantom{-}\cos(\theta) & 0 \\ 0 & 0 & 1 \end{bmatrix} \\
\norm{v} &= \sqrt{\dot{x}^2 + \dot{y}^2} \\
 K_{f} &= \frac{1}{2}\rho C_{f} d_f l_{f}  \left (r_{b}^2 + \frac{1}{4} l_{f}^2 + r_{b} l_{f} \cos(\psi) \right ) \\
 K_{t} &=  r_{b}\cos(\psi) + \frac{1}{2} l_{f} \\
 K_{b} &= \frac{1}{4} \rho C_b A_{sub}
\end{align}
\end{subequations}

All variable definitions are presented in Table \ref{tab:params}, in which the values provided represent the design presented in Section \ref{sec:mechDes}. The values for drag coefficients $C_b$ and $C_f$ are estimated, while $C_r$ is calculated using the model presented for rotating cylinders in \cite{Childs2010}. 

\subsection{Input} \label{sec:input}

As described in Section \ref{sec:model}, when the robot rotates with $\dot{\theta} < 0$ the right flipper is activated, while when $\dot{\theta} > 0$ the left flipper is activated. We can achieve forward motion by alternating left and right flipper paddles, which is accomplished by inputting a periodic function of $\phi$ to force a periodic function of $\theta$.

Following \cite{Refael2018}, the robot we consider is propelled by inputs of the form given in \eqref{eq:input}, which defines a piecewise-continuous sinusoid with varying frequencies $\omega_1$ and $\omega_2$. $T_1$ and $T_2$ are the periods associated with \textit{complete} rotations at frequencies $\omega_1$ and $\omega_2$, respectively, $A$ is the amplitude, and $\phi_0$ is the zero-orientation of the driving mass (the midpoint of the oscillation). The 4-tuple $\left ( T_1 , T_2 , A , \phi_0 \right )$ thus fully defines the input function; the input $(1,1,2,0)$, for example, corresponds to a symmetric cosine wave with period $1\si{s}$, amplitude $2\si{rad}$, and centered around $0 \si{rad}$. 

\begin{equation} \label{eq:input}
    \phi(t) = \begin{cases} 
            \phi_0 + A\cos{\left (\omega_1 t \right )} & t \in \left [0, \frac{T_1}{2} \right )  \\
            \phi_0 - A\cos{\left (\omega_2 \left (t - \frac{1}{2}T_1 \right ) \right )} & t \in \left [\frac{1}{2}T_1,\frac{T_1 + T_2}{2} \right )
        \end{cases}
\end{equation}

\begin{figure}[t]
    \centering
    \includegraphics[width=\linewidth]{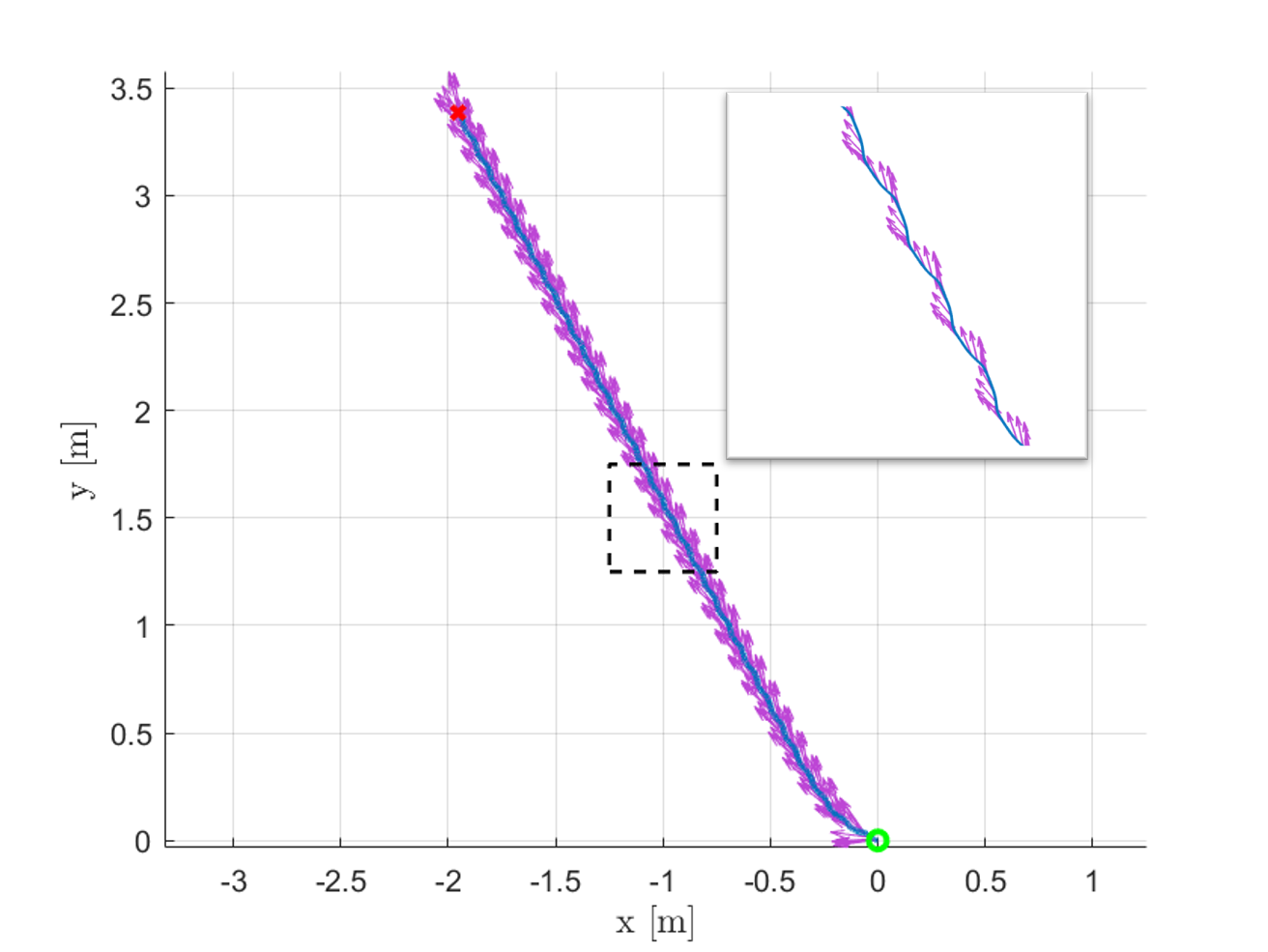}
    \caption{A $30\si{s}$ simulation of the equations of motion \eqref{eq:eom_main} with inputs $(1,1,2,0)$. The boat starts at rest with $\theta (t = 0) = 0$ (upward). The initial location is marked as a green circle, while the final is a red X. The blue line plots the COM, while the arrows indicate the orientation $\theta_t$ (see Section \ref{sec:desChoices}). The inset shows a magnified portion of the trajectory (marked in black) to clarify detail.}
    \label{fig:trajStraight}
\end{figure}

We define a \textbf{stroke} as a period during which the motor rotates in a single direction, i.e. either of the two cases in \eqref{eq:input}. A \textbf{cycle} is then defined as a full period, i.e. two strokes. The robot can then be ``steered'' by varying the periods of oscillation that define the two strokes of the input function $\phi(t)$. When $T_1 = T_2$, the oscillation is symmetric and results in oscillations about a straight line. This can be seen in Fig. \ref{fig:trajStraight}, where numerically solving \eqref{eq:eom_main} with input as in \eqref{eq:input} produces an initial deviation as the robot starts from rest, but then settles into oscillation around a straight line. When $T_1 > T_2$, however, the stroke that activates the right flipper is faster than the stroke that activates the left. Since the angular momentum imparted by the flippers is proportional to $\int \dot{\theta}^2 dt$ (i.e. the drag on a flat plate caused by the rotation of the body in the water), the result is more thrust from the right flipper, which results in a counterclockwise trajectory. When $T_1 < T_2$ the opposite occurs, causing a clockwise trajectory. 

\section{Design for Swarming/Modularity} \label{sec:swarmAndModul}

\subsection{Requirements} \label{sec:requirements}

Although the design presented in Section \ref{sec:basicDes} is innovative in its simplicity, it lacks the ability to move quickly and precisely. Because forward motion is defined by a reciprocal motion of the motor, it is limited by the constant need for the motor to switch directions, so the thrust generated by each stroke is limited.  Moreover, the response time of the system is limited by the fastest cycle-time it can achieve, which defines the lowest amount of time it takes to make a full maneuver. Because of the limited thrust, however, most maneuvers will take longer than one cycle, so the response time is further slowed.

We consider two methodologies for remedying these challenges. The first is to design the swimmer to be \textbf{modular}. The TEMP project \cite{OHara2014}\cite{Paulos2015} showed that modular robot teams made of capable individuals can be useful as structures, but they can also improve mobility when the individual robots are limited, as ours is. Allowing multiple robots to rigidly connect together can potentially allow the group to be more responsive and more powerful, but at a minimum enables more actuated degrees of freedom of the conglomerate. Multiple robots swimming together may provide additional thrust when paddling in phase or more uniform thrust when paddling out of phase. Multiple robots facing in different directions, meanwhile, can make the system more maneuverable by allowing it to turn either more or less effectively or to brake, as the situation requires. 

While the modular approach focuses on overcoming the design's limitations, the second approach --- \textbf{swarming} --- embraces them. In this method, the robots are released en masse. Although no one robot can be relied on to overcome external flows or respond quickly, the group as a whole will be far more resistant to these negative effects. This allows less capable units to perform on par with more functional ones, enabling vast mobile sensor networks or search teams.

\subsection{Design Choices} \label{sec:desChoices}

In order to ensure that our system can function both as a swarming robot or as a modular one, we make design choices that will satisfy the criteria for both methodologies. We therefore make several modifications to the design presented in \cite{Refael2018} when developing a prototype. In particular, we want to ensure that there exists a functional central body with which other robots can interact, whether by docking/undocking or otherwise, and that the motion of the flippers does not interfere with other robots swimming nearby, whether docked or simply in proximity.

\begin{enumerate}
    \item \textbf{Primary Top Body:} we replace the driving mass with a separate, larger body and assign primary focus to its orientation. Thus, while the original configuration included the orientation $\theta$ of the body bearing the flippers, we replace it with the orientation of the new top body $\theta_t$. Because the two bodies are mechanically linked by the motor, the modeling change is achieved by augmenting \eqref{eq:eom_main} with a new orientation variable $\theta_t$ as in \eqref{eq:ori_main}.
    
        \begin{equation} \label{eq:ori_main}
            \theta_t = \theta + \phi
        \end{equation}
    
    \item \textbf{Non-Protruding Flippers:} the flippers are designed such that they do not protrude from the projection of the top body at any time. Assuming the top bodies of neighboring boats are roughly co-planar, top body contact will prevent any collision between flippers. Although it is still possible for flows created by flippers to interfere with the ability of other robots in a group to swim, we do not consider this presently. 
    
    \item \textbf{Docking Points:} magnetic docking points are added at four positions on the top body to allow docking between individual robots. 
    
    \item \textbf{Tail for Undocking:} a feature protruding from the outline of the top body (the ``tail'') is added to the bottom body. This tail --- when suitably designed --- allows the robots to undock from each other by using the existing single actuator.

    \item \textbf{Inexpensive:} the robot must be designed to be inexpensive to allow significant quantities to be produced.
\end{enumerate}

The design created to satisfy these criteria and fit the model presented in Section \ref{sec:basicDes} --- the Modboat --- is presented in Section \ref{sec:mechDes}. 


\section{Mechanical Design} \label{sec:mechDes}

\begin{figure}[t]
    \centering
    \includegraphics[width=\linewidth]{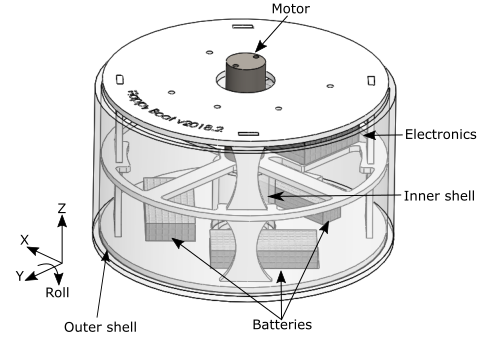}
    \caption{Rendering of the top body of the Modboat. The encoder is mounted on the input shaft of the motor, at the top of the figure, while wiring is not shown for clarity. The coordinate frame has its origin at the center of the body and is fixed to the body.}
    \label{fig:topBody} 
\end{figure}

\begin{figure}[t]
    \centering
    \includegraphics[width=0.85\linewidth]{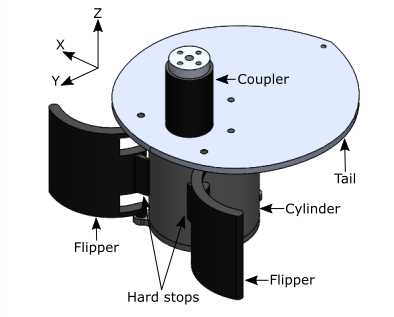}
    \caption{Rendering of the bottom body of the Modboat. The coordinate frame has its origin on the axis of rotation at the base of the coupler, and is fixed to the body.}
    \label{fig:bottomBody}
\end{figure}

The Modboat --- shown in Fig. \ref{fig:renderingIso} --- is comprised of two bodies: (1) a top body that serves as the brain of the robot (shown in Fig. \ref{fig:topBody}), and (2) a bottom body that serves as the propulsion system (shown in Fig. \ref{fig:bottomBody}), mechanically linked by a single DC motor. Both bodies rotate in the water when the robot is operating, so to reduce drag they are shaped as cylinders.

The top body, shown in Fig. \ref{fig:topBody}, is waterproofed and houses five rechargeable NiMH AAA batteries that power the robot. It also contains an ESP8266 microcontroller and motor driver, as well as the motor itself. The point at which the motor shaft exists the body is not waterproofed. Instead, the Modboat is designed to float such that the interface is above the waterline. For these initial single module prototypes the magnetic docking points are omitted.

The top body functions as the driving mass described in Sections \ref{sec:basicDes} and \ref{sec:desChoices}. By including the heaviest components --- the batteries and motor --- in the top body we give it a higher moment of inertia than the bottom body. This allows the Modboat to achieve more rotation from the bottom body using smaller amplitude inputs.

The bottom body, shown in Fig. \ref{fig:bottomBody}, has no actuators or other electronic components. It is comprised of a coupler, which interfaces with the motor, a central cylindrical body, and a tail. The flippers are held in place by loose pin joints on the top and bottom, allowing them to rotate freely between fully closed (against the central cylinder) and fully open (with the hard stops contacting the central cylinder). Their curved shape is designed to lay flat against the central cylinder to maintain a cylindrical profile for reduced drag.

The central cylinder is hollow. This serves to (1) lower the mass and moment of inertia of the bottom body and (2) provide a potential payload compartment. The current bottom body is not waterproofed, but water-safe sensors and payloads may be equipped.

A single 100:1 geared DC motor capable of $14.7 \si{rad/s}$ is used to create relative rotation between the two bodies, with a magnetic encoder to measure orientation. A PID controller run on the ESP8266 ensures that the motor orientation follows prescribed trajectories $\phi(t)$ as defined in \eqref{eq:input}, but the gearbox allows $\approx 5^\circ$ of backlash when static.

The result is an affordable prototype Modboat that costs about $\$122$, as shown in Table \ref{tab:cost}. The components were chosen with only minimal attention to cost; the affordability is rather a result of the simple propulsion mechanism. Once performance is proven, however, the design can be optimized to reduce the cost even further. 

\begin{table}[t]
    \centering
    \caption{Approximate material cost of the Modboat prototype.}
    \begin{tabular}{lccr} \toprule
    Part Name & Cost/Unit (USD) & Qty & Cost (USD) \\ \midrule
    6'' OD acrylic tube & 3.99/inch & 3 & 15.99 \\
    1.25'' acrylic tube & 0.25/inch & 2 & 0.50  \\ 
    12''x12''x1/8'' acrylic & 9.15/sheet & 1 & 9.15 \\
    2'' ABS tube & 1.15/inch & 3 & 3.45 \\ 
    7/8'' ABS rod & 0.55/inch & 3 & 1.65 \\
    12''x12''x1/8'' ABS & 8.87/sheet & 2 & 17.74 \\
    Pololu 20D DC Motor & 22.95 & 1 & 22.95 \\
    Pololu Encoder & 4.48 & 1 & 4.48 \\
    Pololu Mounting Hub & 3.47 & 1 & 3.47 \\
    Flippers & 7.50 & 2 & 15.00 \\ 
    NodeMCU ESP8266 & 8.39 & 1 & 10.99 \\
    Motor Driver {\tiny (TB6612FNG)} & 3.33 & 1 & 3.33 \\
    Custom PCB & 6.20 & 1 & 6.20 \\ 
    Electronic comps. & --- & --- & 4.00 \\ 
    Screws \& Glue & --- & --- & 3.00 \\ \midrule
     & & \textbf{Total:} & 121.90 \\\bottomrule
    \end{tabular}
    \label{tab:cost}
\end{table}


\section{Experiments} \label{sec:experiments}

\begin{figure}[t]
    \centering
    \includegraphics[width=\linewidth]{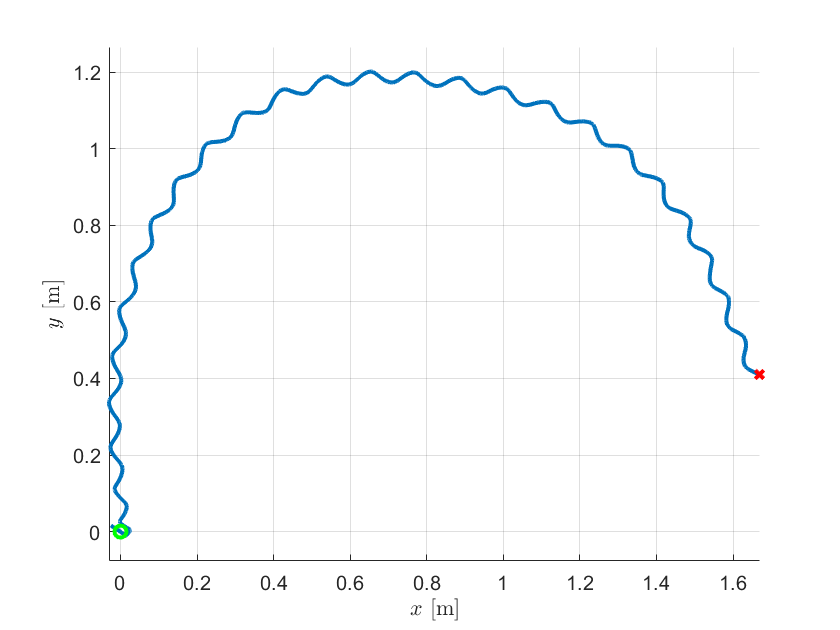}
    \caption{A $30\si{s}$ trajectory of motion capture data from the Modboat running with inputs $(1,1,2,0)$. The initial location is marked as a green circle while the final is a red X, and the trajectory has been artificially rotated so that the initial motion is upwards.}
    \label{fig:trajBad}
\end{figure}

The Modboat was tested in open-loop in a $4.5\si{m} \times 3.0\si{m} \times 1.2 \si{m}$ tank of water equipped with an OptiTrack motion capture system that provided planar position, orientation, velocity, and angular velocity data at $120 \si{Hz}$. A MATLAB interface was used to send motion commands to the Modboat and record the incoming data for post-processing. 

We tested both symmetric and asymmetric input waveforms. The curvature of the resulting trajectories was measured by fitting a circle to the most consistently curved portion of the trajectory and recording its radius, with positive radii assigned to counterclockwise trajectories and negative radii to clockwise trajectories. Ideally, we would see radii approaching infinity for symmetric inputs and finite radii for asymmetric inputs.

\begin{figure}[t]
     \centering
      \subfloat[Data for Modboat with bottom cylinder.\label{fig:weightShiftBottom}]{%
       \includegraphics[width=\linewidth]{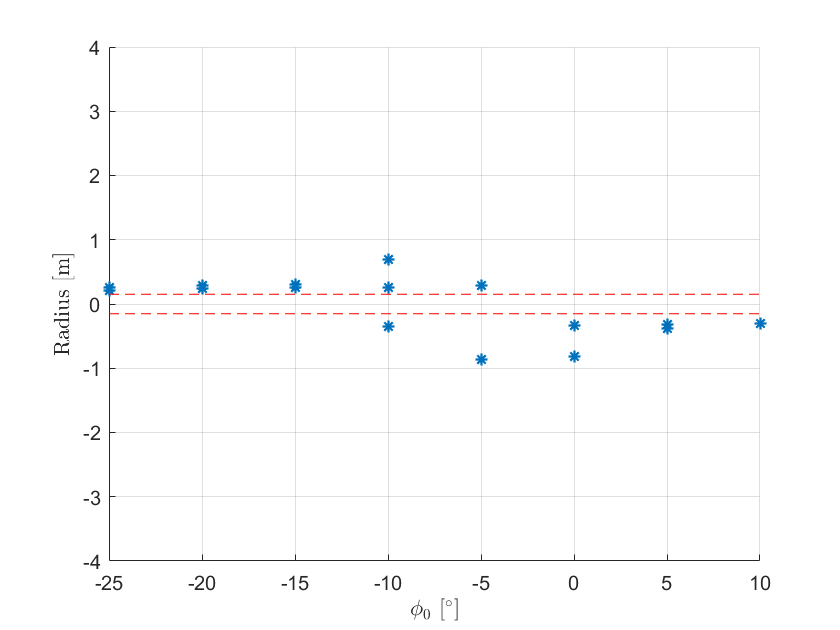}}
       \hfill
      \subfloat[Data for Modboat with $5\si{mm}$ spacers.\label{fig:weightShiftSpacers}]{%
            \includegraphics[width=\linewidth]{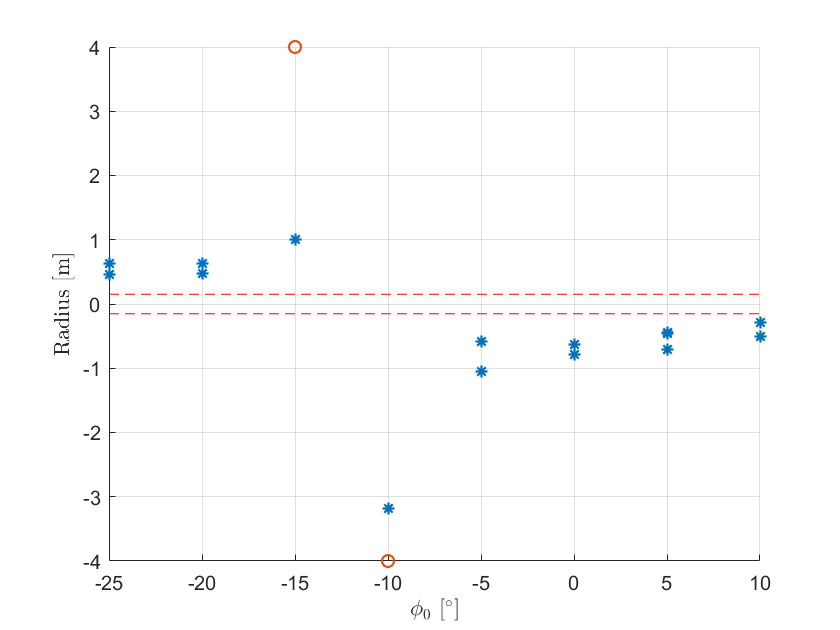}}

        \caption{The radii of the fitted circles plotted against the orientation of the center of mass. The dashed red lines indicate the diameter of the Modboat, while the red circles indicate data truncated to $4\si{m}$ for clarity (since large radii approach straight lines, where the radius ceases to be a useful measure). All tests were conducted for $30\si{s}$ with inputs $(1,1,2,0)$.}
        \label{fig:summary12}
\end{figure}

It is reasonable to expect some deviation from a straight path --- due both to imperfect design of the system and non-static water conditions --- but for symmetric input waveforms we observed strong curves in both clockwise and counterclockwise directions. A sample trajectory is shown in Fig. \ref{fig:trajBad} and demonstrates significant deviation from the straight line trajectory predicted by our model in \eqref{eq:eom_main} for those inputs, such as the one shown in Fig. \ref{fig:trajStraight}.

While there was variation in how much the Modboat turned in a given test, we observed a particularly strong dependence on the location of the center of mass (COM). While the Modboat is designed to align the center of mass with the center of rotation at the motor shaft, some offset exists due to manufacturing imperfections and wiring. If the COM lies along the $y$ axis of the top body (see Fig. \ref{fig:topBody}), then a motion according to \eqref{eq:input} with $\phi_0 = 0$ causes a symmetric oscillation of the COM about the $y$ axis on the bottom body (see Fig. \ref{fig:bottomBody}). If the COM is off-axis, however, it will oscillate asymmetrically about the bottom body axis and may induce curvature in the trajectory. By varying $\phi_0$ in \eqref{eq:input}, we can therefore simulate shifting the COM within the top body and evaluate its influence. 

The radii that result from varying $\phi_0$ are presented in Fig. \ref{fig:weightShiftBottom}, with the diameter of the boat marked for reference. The radii cluster around the diameter lines, indicating sharp turns regardless of center of mass position. We do not observe a region of straight-line behavior, which would be indicated by points tending towards $\pm \infty$.

We found that the on-board motion was symmetric by evaluating the data provided by the encoders, so we considered the effect of the closed position of the flippers on the trajectory curvature. $5\si{mm}$ spacers were placed on the bottom body cylinder to prevent the flippers from fully closing, and the same set of tests was performed. The results are presented in Fig. \ref{fig:weightShiftSpacers}, where we observe that the radii of all the trajectories have increased, and the region around $\phi_0 \in [-15^\circ,-10^\circ]$ now displays curvatures that are much closer to straight lines.


\section{Discussion} \label{sec:discussion}

The sharp turning behavior when the desired trajectories are straight, indicated by the small radii in Fig. \ref{fig:weightShiftBottom}, makes control of this design very difficult. The data indicate that the robot must be precisely balanced to achieve straight line motion; Fig \ref{fig:weightShiftBottom} has only a $5^\circ$ range ($\phi \in [-10^\circ,-5^\circ]$) in which the radii transition from negative to positive, which is where straight lines would occur, despite the fact that we measured the COM as being within $\pm 1.25\si{mm}$ of the motor axis. While achieving sub-millimeter precision in balancing the robot is technically possible, it would (1) increase the cost of the system and (2) mean that we cannot carry payloads or dock with other robots, as these would shift the COM. We could use closed-loop control to improve performance, but these values indicate that there may be a fundamental issue with the design presented causing a heightened sensitivity to mass balance. 

\begin{figure}[t]
    \centering
    \resizebox{\linewidth}{!}{%
    \begin{tikzpicture}

    \coordinate(AX) at (6,8);
    \coordinate(AY) at (13,8);
    
    \draw[gray, thick, ->] (AX) --++ (0,1) node[above] (Y) {$y$};
    \draw[gray, thick, ->] (AX) --++ (1,0) node[right] (X) {$x$};
    
    \path (AY) --++ (0,1) node (Y1) {};
    \path (AY) --++ (1,0) node (X1) {};
    
    \def\radius{1.5}
    \def\angle1{135}
    \coordinate (O) at (10,10);
    \path (O) +(\angle1:\radius) coordinate (FL);
    \path (O) +(45:\radius) coordinate (FR);
    
    \draw[very thick] (O) circle[radius=\radius];
    \fill (O) circle[radius=2pt] node[below right] {O};
    \fill (FL) circle[radius=2pt];
    \fill (FR) circle[radius=2pt];
    
    \begin{scope}[rotate around={60:(FL)}]
    \draw[ultra thick, blue] (FL) arc (0:90:1.5) coordinate (FLEnd);
    \end{scope}
    \draw[dashed, thick] (FL) -- (FLEnd);
    
    \begin{scope}[rotate around={80:(FR)}]
    \draw[ultra thick, blue] (FR) arc (0:-90:1.5) coordinate (FREnd);
    \end{scope}
    
    \draw[dotted] (FL) -- (O);
    
    \begin{scope}[rotate around={90:(FL)}]
    \path (FL) +(\angle1:\radius) coordinate(FLRef);
    \end{scope}
    
    \tkzMarkRightAngle[draw=gray,size=.25](O,FL,FLRef);
    \draw[dotted] (FL) -- (FLRef);
    
    \node[](FLMid) at ($(FL)!0.5!(FLEnd)$) {};
    \fill (FLMid) circle[radius=2pt] node[below right] {};
    \begin{scope}[rotate around={90:(FLMid)}]
    \path (FLMid) + (\angle1:\radius) node[](FLMidRef) {$F_{f}$};
    \end{scope}
    \draw[thick, <-] (FLMid) -- (FLMidRef);
    
    \draw[thick]
        pic["$\alpha$",draw=black, <-, angle eccentricity=1.5, angle radius=.5cm]{angle=FLEnd--FL--FLRef};
    
    \path (O) +(65:1.5*\radius) coordinate (RTD);
    \path (O) +(115:1.5*\radius) coordinate(LTD);
    \draw[dashed, thick]
        pic["$\dot{\theta}$",draw=gray, ->, angle eccentricity=1.2, angle radius=2.0cm]{angle=RTD--O--LTD};;

    \end{tikzpicture}}
    \caption{A simplified model of the Modboat bottom body, showing the flipper opening angle $\alpha$, measured from the perpendicular to the radius. The dashed line represents the flat plate model of the flipper. The drag $F_f$ is shown acting along the tangent line, since we assume in Section \ref{sec:model} that translational velocity is negligible, making the fluid velocity along the tangent.}
    \label{fig:flapOpenAngle}
\end{figure}
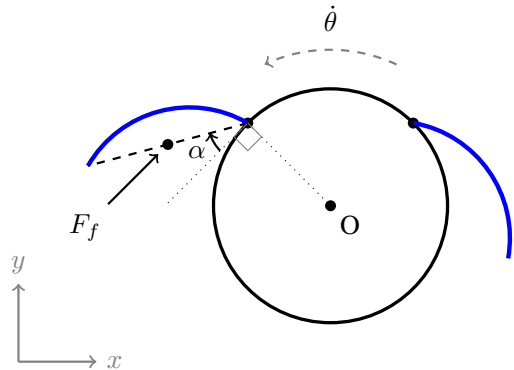

The core of the mechanism for turning and propulsion is the differentiated position of the flippers. Each flipper opens due to two forces: (1) a centrifugal action due to the rotation of the body and (2) drag from the moving water. The drag opening the flipper, shown in Fig. \ref{fig:flapOpenAngle}, acts along the tangent to the circular body, since we assume in Section \ref{sec:model} that the translational velocity of the robot is negligible relative to its rotational velocity. The torque opening the flipper is then given by \eqref{eq:tauOpen}, and is a function of $\sin{(\alpha)}$, where $\alpha$ is the opening angle. This torque produces a positive feedback loop, since (if $\alpha > 0$) as $\alpha$ increases $\sin{(\alpha)}$ increases and so does $\tau_{open}$, further opening the flipper until it is completely open. Under the assumptions made in Section \ref{sec:model}, each flipper produces thrust only when it is fully open, so it is imperative that the flippers open symmetrically when they are driven symmetrically in order to achieve balanced thrust and straight line motion. 

\begin{equation} \label{eq:tauOpen}
    \tau_{open} = \frac{1}{2} l_f F_{f} \sin{(\alpha)}
\end{equation}

When the flippers are completely closed against the bottom body cylinder, very little area is exposed between the flipper and the cylinder and very little water is present in the gap, preventing drag from playing a role. Thus, when the body begins to rotate only the centrifugal action works to open the flipper until water can enter and provide additional drag force. As long as the Modboat floats at zero roll angle when stationary (where roll is the rotation about the body-frame $y$ axis in Fig. \ref{fig:topBody}), this effect is symmetric and the system should conform to the model in \eqref{eq:eom_main}. But we observed that the flippers display a high sensitivity to the roll angle, opening slightly under their own weight at even small non-zero roll angles. The lower flipper --- which floats partially open with water in the gap --- has a potentially significant advantage in opening time and therefore thrust produced. Since even slight offsets of the COM can induce non-zero roll angles in the water, this would cause the Modboat to turn depending on which flipper was given the advantage by this effect.

\begin{figure}[t]
    \centering
    \includegraphics[width=0.9\linewidth]{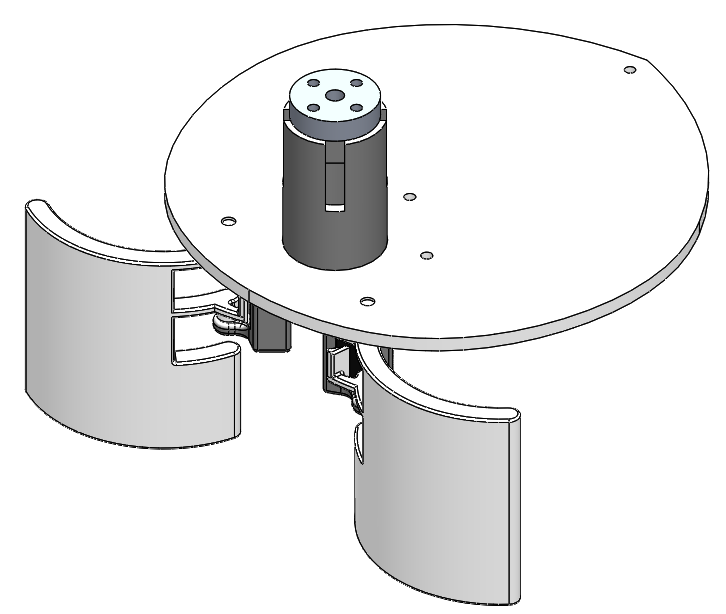}
    \caption{Rendering of the updated bottom body of the Modboat without the bottom body cylinder. This change required a redesign of the flippers to allow a new bottom pin connection, but the overall shape was maintained.}
    \label{fig:bottomBodyNew}
\end{figure}

This explanation would indicate that the weakness of the design presented in Section \ref{sec:mechDes} and shown in Fig. \ref{fig:bottomBody} is the flush closed position of the flippers against the bottom body cylinder. By enforcing a gap between the flippers and the body even when closed, we would allow the flippers to open more symmetrically even when the Modboat has a non-zero roll angle. This is verified in Fig. \ref{fig:weightShiftSpacers}, where $5\si{mm}$ spacers were used to prevent flush closure of the flippers, resulting in larger radii for all trajectories and significantly larger radii around $\phi_0 \in [-15^\circ,-10^\circ]$, which now are beginning to approach straight lines. This indicates that drag is the dominant effect in determining how symmetrically the flippers open, and that allowing it to act is critical. Nevertheless, while the $5\si{mm}$ spacers are an improvement, the drop in radius for $\phi_0 > -10^\circ$ and $\phi_0 < -15^\circ$ shows that the COM dependence is still present and further improvement is needed.

\begin{figure}[t]
    \centering
    \includegraphics[width=\linewidth]{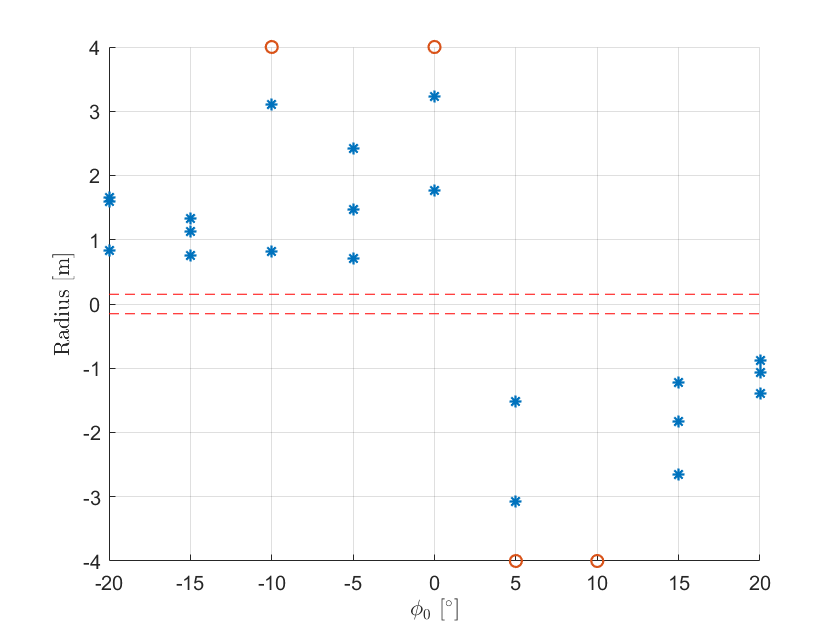}
    \caption{The radii of the fitted circles plotted against against the orientation of the center of mass for the Modboat with \textbf{no bottom cylinder}. The dashed red lines indicate the diameter of the Modboat, and the red circles indicated truncated data as in Fig. \ref{fig:weightShiftSpacers}. All tests were conducted for $30 \si{s}$ with inputs $(1,1,2,0)$.}
    \label{fig:weightShiftNoBottom}
\end{figure}

A redesigned bottom body that removes the cylinder entirely is shown in Fig. \ref{fig:bottomBodyNew}; the flippers have been modified slightly to continue to allow a two pin connection, but their overall shape has been maintained. This design allows water to access the flippers at all times so that drag can act immediately when they begin to open, and this reduces the dependence on the roll angle significantly. Fig. \ref{fig:weightShiftNoBottom} shows the results for testing this new design, where we observe that the sharpest turns are in the same region as the widest turns of Fig. \ref{fig:weightShiftBottom} and we can achieve relatively large radii over a much wider range of COM offsets. While these results are still far from ideal, they are effective enough to allow closed-loop control to finish the job.

Although we have removed the payload compartment from this new bottom body, we can still carry water-safe payloads as long as they do not prevent water from reaching the flippers. Additionally, a payload compartment can still be designed as long as it meets this restriction. The ability to dock with other robots for modularity has also not been restricted by the redesign, as the flippers are still contained within the top body profile.


\section{Conclusions}

In this paper we have presented and characterized a new design for an affordable single-motor swimming robot --- the Modboat --- based on \cite{Refael2018}, which is unique in allowing modular and swarming behaviors. These behaviors are aimed to overcome the main limitations of the design, which are its low thrust and maneuverability, while building on its strengths: simplicity and inherent low cost. 

We have shown experimentally that this robot is subject to significant sensitivity to the position of its center of mass, effectively removing its ability to swim in straight lines. This is most likely caused by the non-zero roll angle that offset mass induces on the robot, which accentuates asymmetries in flipper thrusts. 

Any robot implementing this design must be built to reduce this sensitivity  in order to provide reasonable open-loop behavior that can be stabilized in closed-loop. We have shown that redesigning the passive bottom body of the Modboat significantly improves the open-loop performance of the system in attempting to swim straight lines. The resulting trajectories are considered reasonable enough to allow closed-loop control to stabilize the system.

In future work we will to implement closed-loop orientation feedback and evaluate the ability of the Modboat to follow trajectories and perform tasks using this methodology. This type of position control will be critical to enable the docking of multiple boats together as we explore swarming and larger group behaviour. The impact of disturbances, such as flows or vortices, will also be considered, as will different motion primitives that may be more successful at driving the system.


\section*{Acknowledgment}

We thank Dr. M. Ani Hsieh for graciously allowing us the use of her tank and motion capture system for all of the testing and data collection described in this work.


\bibliographystyle{./bibliography/IEEEtran}
\bibliography{./bibliography/IEEEabrv,./bibliography/ur2020}

\begin{thebibliography}{10}
\providecommand{\url}[1]{#1}
\csname url@rmstyle\endcsname
\providecommand{\newblock}{\relax}
\providecommand{\bibinfo}[2]{#2}
\providecommand\BIBentrySTDinterwordspacing{\spaceskip=0pt\relax}
\providecommand\BIBentryALTinterwordstretchfactor{4}
\providecommand\BIBentryALTinterwordspacing{\spaceskip=\fontdimen2\font plus
\BIBentryALTinterwordstretchfactor\fontdimen3\font minus
  \fontdimen4\font\relax}
\providecommand\BIBforeignlanguage[2]{{%
\expandafter\ifx\csname l@#1\endcsname\relax
\typeout{** WARNING: IEEEtran.bst: No hyphenation pattern has been}%
\typeout{** loaded for the language `#1'. Using the pattern for}%
\typeout{** the default language instead.}%
\else
\language=\csname l@#1\endcsname
\fi
#2}}

\bibitem{Leonard2007}
\BIBentryALTinterwordspacing
N.~E. Leonard, D.~A. Paley, F.~Lekien, R.~Sepulchre, D.~M. Fratantoni, and
  R.~E. Davis, ``{Collective motion, sensor networks, and ocean sampling},''
  \emph{Proceedings of the IEEE}, vol.~95, no.~1, pp. 48--74, jan 2007.
  [Online]. Available: \url{http://ieeexplore.ieee.org/document/4118466/}
\BIBentrySTDinterwordspacing

\bibitem{Leonard2010}
\BIBentryALTinterwordspacing
N.~E. Leonard, D.~A. Paley, R.~E. Davis, D.~M. Fratantoni, F.~Lekien, and
  F.~Zhang, ``{Coordinated control of an underwater glider fleet in an adaptive
  ocean sampling field experiment in Monterey Bay},'' \emph{Journal of Field
  Robotics}, vol.~27, no.~6, pp. 718--740, nov 2010. [Online]. Available:
  \url{http://doi.wiley.com/10.1002/rob.20366}
\BIBentrySTDinterwordspacing

\bibitem{Mintchev2014}
S.~Mintchev, E.~Donati, S.~Marrazza, and C.~Stefanini, ``{Mechatronic design of
  a miniature underwater robot for swarm operations},'' in \emph{Proceedings -
  IEEE International Conference on Robotics and Automation}.\hskip 1em plus
  0.5em minus 0.4em\relax Institute of Electrical and Electronics Engineers
  Inc., sep 2014, pp. 2938--2943.

\bibitem{OHara2014}
I.~O'Hara, J.~Paulos, J.~Davey, N.~Eckenstein, N.~Doshi, T.~Tosun, J.~Greco,
  J.~Seo, M.~Turpin, V.~Kumar, and M.~Yim, ``{Self-assembly of a swarm of
  autonomous boats into floating structures},'' in \emph{Proceedings - IEEE
  International Conference on Robotics and Automation}.\hskip 1em plus 0.5em
  minus 0.4em\relax Institute of Electrical and Electronics Engineers Inc., sep
  2014, pp. 1234--1240.

\bibitem{Paulos2015}
J.~Paulos, N.~Eckenstein, T.~Tosun, J.~Seo, J.~Davey, J.~Greco, V.~Kumar, and
  M.~Yim, ``{Automated Self-Assembly of Large Maritime Structures by a Team of
  Robotic Boats},'' \emph{IEEE Transactions on Automation Science and
  Engineering}, vol.~12, no.~3, pp. 958--968, jul 2015.

\bibitem{Furno2017}
L.~Furno, M.~Blanke, R.~Galeazzi, and D.~J. Christensen,
  ``{Self-reconfiguration of modular underwater robots using an energy
  heuristic},'' in \emph{IEEE International Conference on Intelligent Robots
  and Systems}, vol. 2017-Septe.\hskip 1em plus 0.5em minus 0.4em\relax
  Institute of Electrical and Electronics Engineers Inc., dec 2017, pp.
  6277--6284.

\bibitem{Mintchev2014Journal}
S.~Mintchev, R.~Ranzani, F.~Fabiani, and C.~Stefanini, ``{Towards docking for
  small scale underwater robots},'' \emph{Autonomous Robots}, vol.~38, no.~3,
  pp. 283--299, 2014.

\bibitem{Kelly2012}
\BIBentryALTinterwordspacing
S.~D. Kelly, M.~J. Fairchild, P.~M. Hassing, and P.~Tallapragada,
  ``{Proportional heading control for planar navigation: The Chaplygin beanie
  and fishlike robotic swimming},'' \emph{Proc. American Control Conference},
  pp. 4885--4890, 2012. [Online]. Available:
  \url{http://ieeexplore.ieee.org/xpls/abs{\_}all.jsp?arnumber=6315688}
\BIBentrySTDinterwordspacing

\bibitem{Degani2010}
A.~Degani, H.~Choset, and M.~T. Mason, ``{DSAC - Dynamic, Single Actuated
  Climber: Local stability and bifurcations},'' \emph{Proceedings - IEEE
  International Conference on Robotics and Automation}, pp. 2803--2809, 2010.

\bibitem{Degani2016}
A.~Degani, ``{Dynamic single actuator robot climbing a chute: Period-doubling
  bifurcations: analysis and experiments},'' \emph{Meccanica}, vol.~51, no.~5,
  pp. 1227--1243, 2016.

\bibitem{Tallapragada2015}
P.~Tallapragada, ``{A swimming robot with an internal rotor as a nonholonomic
  system},'' \emph{Proceedings of the American Control Conference}, vol.
  2015-July, pp. 657--662, 2015.

\bibitem{Refael2015}
G.~Refael and A.~Degani, ``{Momentum-driven single-actuated swimming robot},''
  \emph{IEEE International Conference on Intelligent Robots and Systems}, vol.
  2015-Decem, pp. 2285--2290, 2015.

\bibitem{Refael2018}
------, ``{A Single-Actuated Swimming Robot: Design, Modelling, and
  Experiments},'' \emph{Journal of Intelligent and Robotic Systems: Theory and
  Applications}, pp. 1--19, 2018.

\bibitem{Childs2010}
P.~R.~N. Childs, ``{Rotating Cylinders, Annuli, and Spheres},'' in
  \emph{Rotating Flow}.\hskip 1em plus 0.5em minus 0.4em\relax Elsevier, 2010,
  ch.~6, pp. 177 -- 247.

\end{thebibliography}


\end{document}